\documentclass[letterpaper, 10 pt, conference]{ieeeconf}  

\IEEEoverridecommandlockouts                              

\usepackage{mathptmx}
\usepackage{algorithmic}
\usepackage{graphicx} 
\usepackage{amsmath}
\usepackage{amssymb} 
\usepackage{booktabs}
\usepackage{soul}
\usepackage{algorithm}
\usepackage{xcolor}

\usepackage{amsthm}
\usepackage{hyperref}
\usepackage{tabularx}

\title{\LARGE \bf
Flexible Handover with Real-Time Robust Dynamic Grasp Trajectory Generation
}

\author{Gu Zhang$^{1}$, Hao-Shu Fang$^{1}$, Hongjie Fang$^{1}$, Cewu Lu$^{2}$
\thanks{$^{1}$Gu Zhang, Hao-Shu Fang and Hongjie Fang are with the Department of
Computer Science, Shanghai Jiao Tong University, Shanghai 200240, China (e-mail: blake-nash@sjtu.edu.cn; fhaoshu@gmail.com; galaxies@sjtu.edu.cn).
        }%
\thanks{$^{2}$Cewu Lu is the corresponding author, member of Qing Yuan Research
Institute and MoE Key Lab of Artificial Intelligence, AI Institute, Shanghai
Jiao Tong University, Shanghai 200240, China (e-mail: lucewu@sjtu.edu.cn).
        }%
}

\begin{document}

\maketitle
\thispagestyle{empty}
\pagestyle{empty}

\begin{abstract}
    In recent years, there has been a significant effort dedicated to developing efficient, robust, and general human-to-robot handover systems. However, the area of flexible handover in the context of complex and continuous objects' motion remains relatively unexplored. In this work, we propose an approach for effective and robust flexible handover, which enables the robot to grasp moving objects with flexible motion trajectories with a high success rate. The key innovation of our approach is the generation of real-time robust grasp trajectories. We also design a future grasp prediction algorithm to enhance the system's adaptability to dynamic handover scenes. We conduct one-motion handover experiments and motion-continuous handover experiments on our novel benchmark that includes 31 diverse household objects. The system we have developed allows users to move and rotate objects in their hands within a relatively large range. The success rate of the robot grasping such moving objects is 78.15\% over the entire household object benchmark.

\end{abstract}

\section{Introduction}

Human-robot collaboration has gained considerable attention in the robotics community. The ability to successfully and reliably transfer objects from humans to robots is a crucial component in enabling robots to assist with manipulation tasks in a flexible and effective manner~\cite{ortenzi2021object}. Building a reactive, general, and robust autonomous handover system is a major challenge for researchers in the field.

Numerous methods have been proposed to develop effective human-to-robot handover systems, including intention cognition, motion control, and trajectory planning. Advances in general object grasping strategies have significantly enhanced the generalizability of handover~\cite{rosenberger2020object, yang2021reactive}. However, most existing methods require that the object being handed over be relatively stationary after the robot begins to move, which greatly limits the flexibility of the system and its potential to handle more complex collaborative tasks. We refer to this type of handover as static handover. The opposite type is dynamic handover, which means that the object can still be in motion after the robot begins to move. While some literature attempts to address dynamic handover~\cite{yang2021reactive, marturi2019dynamic, mavsar2022rovernet}, these methods require an external command to execute final grasping or only allow for simple motion patterns. By contrast, we aim to build a closed-loop autonomous system that can successfully perform handover even when the object being handed over exhibits complex movements within a relatively large range during the entire process. Such handover places minimal restrictions on the type, motion pattern, and movement range of the moving objects to be handed over. Hence, we define it as the flexible handover, which can greatly enhance the  adaptability of human-robot interaction.
\begin{figure}[t!]
    \centering
    \includegraphics[scale=0.48]{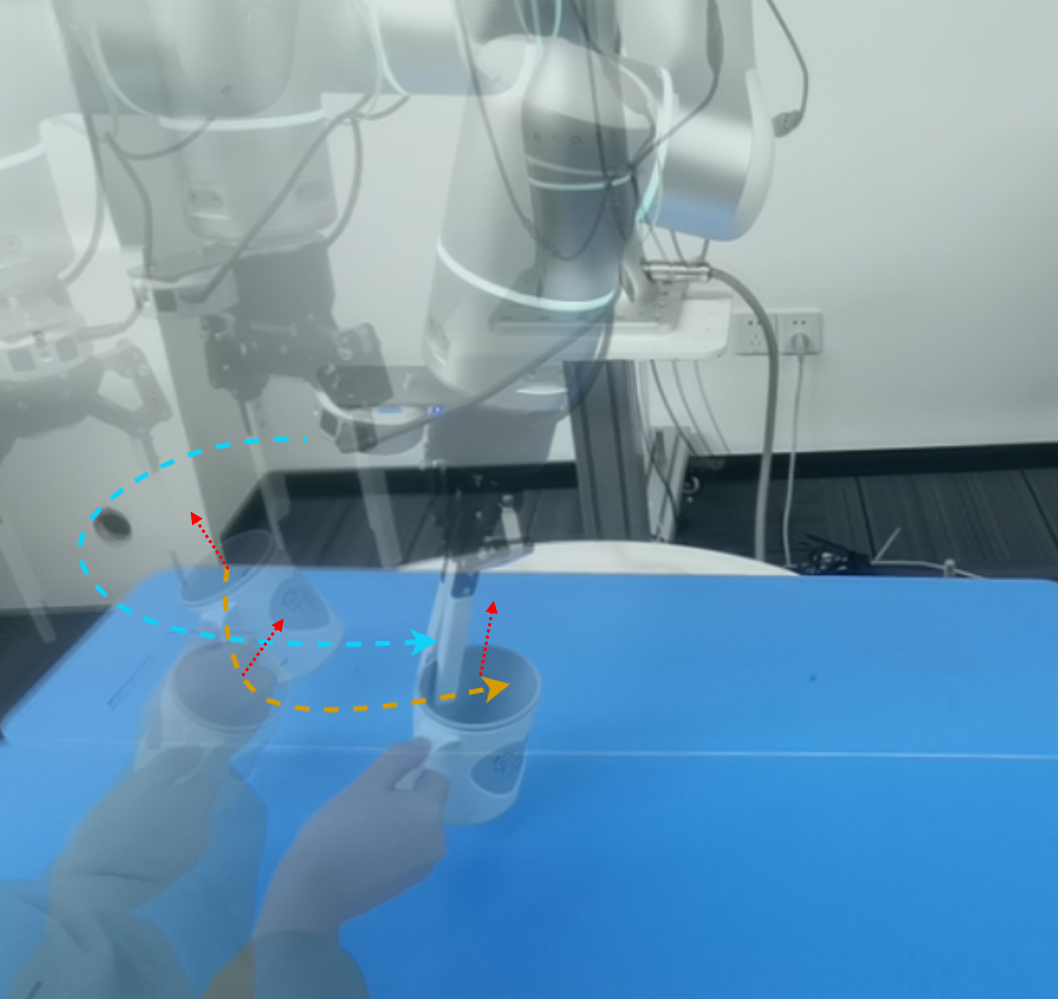}
    \caption{Illustration of flexible handover. Yellow and blue dashed lines denote the moving trajectory of the target object and a robot arm respectively. Red arrows denote the variation in the object's orientation. Our approach enables the human-to-robot handover successful under the flexible and continuous motion of objects. Our system also generalizes well over arbitrary objects. More experiments are conducted over our novel household object benchmark with 31 diverse objects. }
    \label{fig:motion}
\end{figure}

In this paper, we present a novel approach to achieving a flexible handover system. First, we utilize GSNet~\cite{Wang_2021_ICCV}, a general object grasping framework, to guarantee the generalization ability of the handover system. Then building upon the grasp detection backbone, we propose an efficient method for generating real-time robust dynamic grasp trajectories that enhance the stability of the robot's motion trajectories and maintain consistent grasp quality. Additionally, we have designed a future grasp prediction algorithm that leverages the tracking trajectory to predict the future pose, further improving the success rate of flexible handover.

We construct a benchmark consisting of 31 household objects to assess the effectiveness of our handover system under flexible dynamic scenes. To evaluate the performance of our system, we conduct two types of experiments using this benchmark:
\begin{itemize}
    \item \textbf{One-motion}. In the experiment, the object is moved only once after the robot starts moving following different patterns (rotation or translation). 6 objects are selected from the benchmark to measure the approaching time and success rate.
    \item \textbf{Motion-continuous}. The experiment participant can move and rotate objects to be handed flexibly within a wide range during the entire handover process. We test the success rate of the proposed flexible handover system under such random, complex, and continuous motion over the entire benchmark.
\end{itemize}

Evaluation results have shown the reactivity and effectiveness of our flexible handover system.

\section{Related Works}
Researchers have conducted studies on human-to-robot handover from various perspectives. Some researchers focused on how to recognize the human intentions for controlling the object during handover. They used vision-based methods~\cite{strabala2013toward, aleotti2012comfortable, grigore2013joint} or physical-based methods~\cite{nagata1998delivery, edsinger2007human, wang2018controlling} (\textit{e.g.} tactile, force-torque, or wearing sensing) to detect and localize the human's intention, motion and robot's behavior during the object's delivery between human and robot. Another key point is motion planning and robot control for handover. There is a lot of work for human-robot handover to learn effective strategies from human-human handovers. Kajikawa \textit{et al.} planned the robot's motion from the analysis results of trajectories and velocity patterns during a human-human handover~\cite{kajikawa2000trajectory}. Vogt \textit{et al.} used an imitation learning methodology to allow the robot to learn an interaction model from human-human handovers which can generalize the demonstration to new conditions~\cite{vogt2018one}. 
Some other research works utilized Model Predictive Control (MPC) or Dynamic Movement Primitives (DMP) to control robots for smoother motion during handover~\cite{yang2022model,prada2014implementation}. Aside from the vision-based handover systems, researchers also explored autonomous object handover using other information like tactile or force sensing~\cite{bianchi2018touch,konstantinova2017autonomous,mazhitov2023human}.

Recently, with the development of technology in grasp pose detection, human-to-robot handovers became more general, and unseen objects can be delivered and received successfully. Rosenberger \textit{et al.}~\cite{rosenberger2020object} utilized a generic object detector YOLO v3~\cite{redmon2018yolov3} and an effective grasp detection algorithm GG-CNN~\cite{morrison2018closing} to develop an approach for object-independent human-to-robot handovers.~\cite{yang2021reactive} enabled reactive human-to-robot handovers of unknown objects using a model-free grasp pose detector 6-DoF GraspNet~\cite{mousavian20196}. 

On the basis of a general handover system, enabling the robot to grasp moving objects during a handover can greatly enhance the flexibility, reactivity, and robustness of the human-to-robot handover. Marturi \textit{et al.} used the particle filter-based 3D tracker~\cite{aitor2012point} to track the pose of moving object and adjusted generated grasp trajectories to finally select a collision-free and steady grasp trajectory by evaluation online~\cite{marturi2019dynamic}. However, the system focused on the operation of fixed grasp trajectory and required the external signal to allow the grasp execution. The permitted changes in the pose of the moving objects are also limited. Yang \textit{et al.}~\cite{yang2021reactive} proposed an approach to generate temporally consistent grasp by evaluating perturbed grasp candidates. It is not effective with flexible motion and doesn't guarantee the consistency of grasp quality. In AnyGrasp~\cite{fang2023anygrasp}, a temporal association module was designed to learn grasp correspondences between two consecutive frames. Building upon this, Liu et al.~\cite{Liu_2023_CVPR} proposed a refinement module that utilizes historical information to adjust the grasp pose.~\cite{fang2023anygrasp} and~\cite{Liu_2023_CVPR} both focus on "multi-grasp tracking" instead of directly outputting a stable grasp trajectory. However, these approaches are not straightforward, and the complex network architectures can potentially slow down inference speed and reduce success rates. Mavsar \textit{et al.}~\cite{mavsar2022rovernet} utilized a human-body pose estimation algorithm to track a human's hand and predict the best handover 3D position based on the trajectory. This approach lacks generalization for types of different objects and has significant restrictions on the pose of the grasped object.

As to the evaluation of dynamic handover, Yang \textit{et al.} performed a systematic evaluation to record time and success rate over three different objects with different rotations~\cite{yang2021reactive}, where the object was moved only once after the robot began to move, which we call as one-motion handover experiments. Wang \textit{et al.}~\cite{wang2022goal} updated the grasping policy in~\cite{yang2021reactive} and tested the success rate of dynamic handover, where the object was translated from one side to another during the process. The motion pattern in the experiment is still relatively simple. In contrast to these, we conduct experiments where the moving objects can have complex motion patterns and a wide movement range, which we refer to as motion-continuous experiments. A high success rate demonstrated in such experiments can provide evidence that the implemented handover system meets the requirements for flexible handover.

\section{Methods}
In this section, we first introduce the preliminaries of the grasp detection backbone that we adopt. We then describe our tracking network which is used to generate real-time robust dynamic grasp trajectories, as well as the algorithm used to predict future grasps. 

\subsection{Grasp Detection}
GSNet~\cite{Wang_2021_ICCV} is a universal grasping framework that can generate accurate and dense 7-DoF grasp poses given the point cloud of the cluttered scene and achieve a high inference speed. Every detected grasp pose $G$ is represented by [$\mathbf{R}$, $\mathbf{t}$, $w$], where $\mathbf{R}\in \mathbb{R}^{3 \times 3}$ denotes the rotation of grasp $G$, $\mathbf{t} \in \mathbb{R}^{3 \times 1}$ denotes the translation of grasp $G$ and $w \in \mathbb{R}$ is the width of gripper required for grasping the object. 

One core contribution of GSNet~\cite{Wang_2021_ICCV} is the introduction of ``graspness'', a quality that discriminates graspable area to enhance grasp sampling. GSNet~\cite{Wang_2021_ICCV} consists of two-stage models: cascaded graspness model which uses the Minkowski-ResUNet~\cite{diakogiannis2020resunet},~\cite{choy20194d} backbone and MLP blocks to generate point-wise graspable landscape (graspable FPS) and view-wise graspable landscape (graspable PVS) given dense point cloud, and grasp operation model to predict translation, rotation, width, and score of ${M}$ grasp poses. 

The entire architecture is end-to-end, utilizing the point cloud of a cluttered scene to directly predict graspable poses. GSNet~\cite{Wang_2021_ICCV} has a strong generalization ability for grasping unseen objects and runs efficiently. This is why we
opt to use GSNet~\cite{Wang_2021_ICCV} as the grasping detection backbone. In practice, we further reduce the number of approach directions to half of what was used in the original paper~\cite{Wang_2021_ICCV} and simplify the backbone to speed up the detection.

\subsection{Real-Time Robust Dynamic Grasp Trajectory Generation}
The original GSNet~\cite{Wang_2021_ICCV} is designed for static scenes. However, in the dynamic handover experiments, if the robot attempts to approach the pose based solely on the highest grasp score per frame, the resulting trajectory can become erratic~\cite{yang2021reactive}. Therefore, how to generate a smooth, stable, and robust dynamic grasp trajectory in real-time is one of the key points for conducting a safe and effective dynamic handover. 

A stable and smooth grasp trajectory should ensure the convergence of the target approaching pose during tracking. It is also acknowledged that the quality of a grasp pose is determined by its relative position to the target object. Therefore, a grasp trajectory that guarantees all its poses have a close SE(3) distance in the object coordinate frame can not only ensure robustness but also maintain the consistency of grasp quality. To achieve this, We develop a network based on a lightweight transformer~\cite{vaswani2017attention} to learn such property to track the approaching pose in real-time for the current frame utilizing history information of past $T-1$ frames. The resulting tracking poses comprise the smooth and robust grasp trajectory. 

\subsubsection{Dataset} We use GraspNet-1Billion~\cite{fang2020graspnet} for training our tracking network. This dataset offers a diverse set of cluttered scenes, along with abundant information from 256 different camera viewpoints of each scene, such as RGB, depth, object poses, segmentation, and more. For each viewpoint $\mathbf{v_i}$, we denote $\mathbf{t_{\text{Cam}_i}} \in \mathbb{R}^{3 \times 1}$ as the translation of corresponding camera pose. We then apply sphere search to randomly select $T-1$ other viewpoints $\mathbf{v_j}$ satisfying $\|\mathbf{t_{\text{Cam}_i}}-\mathbf{t_{\text{Cam}_j}}\|< 0.1\text{m}$.  This approach enables us to obtain 256 time slices of $T$ frames $\{F_j\}_{j=0}^{T-1}$ for each cluttered scene. Our tracking network is trained over every $T$ frames.

Based on the original camera motion, we apply data augmentation in camera coordinates to point clouds for better real-world motion simulation. For each time slice $\{F_j\}_{j=0}^{T-1}$, we select the grasp pose $\hat G_0$ with the highest score in the first frame $F_0$. Then we calculate the distances of grasp candidates in another frame $F_j$ relative to $\hat G_{0}$ in the object's coordinate, and the one with the nearest distance is taken as the ground truth $\hat G_j$ in frame $F_j$. 

\begin{figure}[htp]
    \centering

    \includegraphics[scale=0.5]{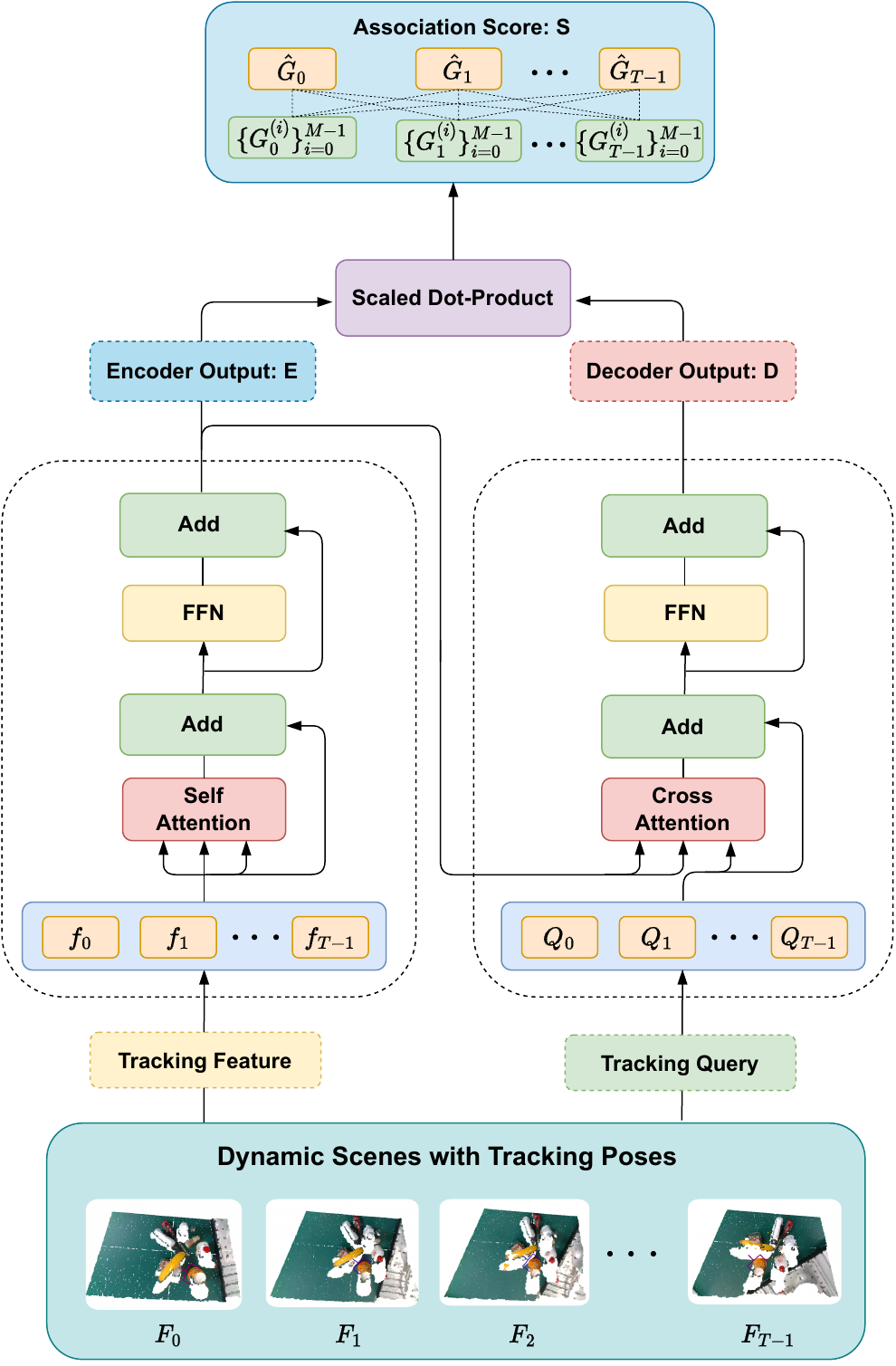}
    \caption{The architecture of tracking network. The scaled dot-product is defined as $S=DE^\top/\sqrt{C}$ where $C$ is the feature dimension.}
     \label{fig:track}
\end{figure}

\subsubsection{Network Architecture} GSNet~\cite{Wang_2021_ICCV} samples $M$ seed points and outputs the grasp pose with the highest score for each seed point. The number of detected poses for each scene is fixed. Therefore, we use $G^{(i)}_j, i \in $ \{${0, {1}, ..., {M-1}}$\} to represent the $i$-th detected pose in frame $F_j$, and $\hat G_j$ to denote the ground truth of tracking pose.

Our first step is to extract features for every detected grasp pose. We utilize and adapt the extraction method presented in~\cite{fang2023anygrasp}. We obtain the geometric features of seed points by combining the point features extracted from the output of GSNet's backbone and the view features which are the input of the last layer of the MLP block to generate PVS in the GSNet.  Similarly, grasp features are extracted from the last MLP block of the grasp operation model in the GSNet. All the above features are obtained from the input point cloud. RGB information can enhance the representation capability of a grasp detector~\cite{gou2021RGB}, hence we sample the color information along the approaching vector within a cylindrical region centered around the seed point, and then use MLP to get the color features~\cite{fang2020graspnet,Wang_2021_ICCV}. Geometric features, grasp features, color features, and grasp pose information are then concatenated together into the tracking features for each grasp candidate.

The grasping tracking network architecture is adapted from GTR~\cite{zhou2022global}, a light-weighted and effective network to solve multi-object tracking in the computer vision field. We make the following modifications:
\begin{itemize}
    \item We replace the ordinary dot-product with a scaled dot-product to better facilitate the convergence of the transformer training;
    \item We replace ReLU~\cite{glorot2011deep} activation function to GeLU~\cite{hendrycks2016bridging} which is more commonly used in transformers;
\end{itemize}

As illustrated in Fig.~\ref{fig:track}, the tracking features $f_j \in \mathbb{R}^{M \times C} $ over the whole time window $\{F_j\}_{j=0}^{T-1}$ are concatenated together into $f^* \in \mathbb{R}^{N \times C} $ as the input of the encoder, where $N = T\times M$. $Q_t \in \mathbb{R} ^C$ is the query for the ground truth tracking pose in the $t$-th frame, where the query is set to be the tracking feature of the ground truth pose. Similar to $f^*$, $Q_t$ from different frames are concatenated into $Q^* \in \mathbb{R}^{T \times C}$ as the input of the decoder. Through the transformer, the scaled dot product is conducted between the output of the encoder $E \in  \mathbb{R}^{N \times C}$ and the output of the decoder $D \in  \mathbb{R}^{T \times C}$ to generate the association score $S \in \mathbb{R} ^{T\times N}$ between tracking queries $Q_t$ and all the pose candidates across $T$ frames. 

\subsubsection{Training and Inference} We calculate the probability distribution of the ground truth tracking poses in $F_j$ induced by $Q_t$ using softmax as follows:
\begin{equation}
\label{eq:prob}
    P(G_j^{(i)}=\hat G_j \mid Q_t)=\frac {\exp(S_{t,jM+i})}{\sum_{k=0}^{M-1}\exp(S_{t,jM+k})}
\end{equation}
As GSNet~\cite{Wang_2021_ICCV} can detect dense grasp poses, the grasp poses with a small distance to the ground truth in the object's coordinate can also be considered as ground truth. Therefore, we use cross-entropy-based tolerance loss $L_\text{tolerance}$ for training.
\begin{equation}
    L_\text{tolerance} = -\sum_{t=0}^{T-1} \sum_{j=0}^{T-1} \frac{1}{n_j}\sum_i^{d(\tilde{G}_j^{(i)},\hat{G}_j)<\tau} \log P(\tilde{G}_j^{(i)}=\hat G_j|Q_t)
\end{equation}
where $\tau = 0.01\text{m}$ is the tolerance threshold, and $n_j$ is the number of tolerance grasp poses $\tilde{G}_j^{(i)}$.

During the inference, the $T-1$ history frames serve as the frames from $F_0$ to $F_{T-2}$. We calculate the average probability for the current frame $F_{T-1}$ to select the tracking pose with maximum association probability after filtering, as shown in Eqn.~\eqref{eq:inference}. If there are less than $T$ history frames, we can similarly calculate the tracking pose using existing frames, since the transformer is flexible to the length of the input.
\begin{equation}
\label{eq:inference}
    P(G_{T-1}^{(i)}=\hat G_{T-1}) = \frac{1}{T-1}\sum_{t=0}^{T-2} P(G_{T-1}^{(i)}=\hat G_{T-1}|Q_t)
\end{equation}

\subsubsection{Implementation Details} The number of frames $T$ is set to 3 for a combination of performance and speed. As to the data augmentation, we randomly rotate the point cloud by $(\alpha,\beta,\gamma)$ along $x$-, $y$-, and $z$-axis respectively, where $\alpha,\beta\sim U\left[-22.5^{\circ}, 22.5^{\circ}\right]$ and $\gamma\sim U\left[-30^{\circ}, 30^{\circ}\right]$; and randomly translate the point cloud by $(\Delta x, \Delta y, \Delta z)$ along $x$-, $y$- and $z$-axis respectively where $\Delta x, \Delta y \sim U\left[-0.2\text{m}, 0.2\text{m}\right]$ and $\Delta z\sim U\left[-0.25\text{m}, 0.25\text{m}\right]$. For the network, we employ dropout~\cite{srivastava2014dropout} with probability $p = 0.1$ to the output after attention.
During training, the parameters of the grasp detection module are frozen. We use Adam~\cite{kingma2014adam} as the optimizer and train the network for 20 epochs. The initial learning rate is set to 0.0001 and weight decay is not applied. The learning rate scheduler follows the polynomial policy employed in~\cite{chen2017deeplab}, where the power index is set to 0.9. The network is trained on one NVIDIA RTX 2080Ti GPU with a batch size of 4. 

\subsection{Future Grasp Prediction}
\label{sec:prediction}

\begin{algorithm}[t]

\caption{Future Grasp Prediction}
\label{algo:alg1}
\textbf{Input}: tracking pose $\hat G_j$ in $F_j$

\textbf{Output}: predicted future grasp pose $\bar {G_j}$

\begin{algorithmic}[1]

\STATE Append $\hat{G}_j$ to the grasp list $L$
\STATE ${\Delta L} \leftarrow$  Diff($L$)
\STATE Select a relative stable sequence ${\Delta L_s}$ from $\Delta L$, where $\Delta L_s^i = (\Delta t_x^i,\Delta t_y^i, \Delta t_z^i)$ satisfying $\Delta t_{x,y,z}^i< \delta_s$. 
\STATE $\Delta T_{x_+} \leftarrow  \{{\Delta t_{x}^i}\mid {\Delta t_{x}^i}>-\delta_p\}$
\STATE $\Delta T_{x_-} \leftarrow  \{{\Delta t_{x}^i} \mid {\Delta t_{x}^i}<\delta_p\}$
\IF{$|\Delta T_{x_+}| \geq |\Delta T_{x_-}|$}{ 
\STATE $\Delta \bar{t}_x \leftarrow |\Delta T_{x_+}|^{-1}\sum_{t_x \in \Delta T_{x_+}} t_x$
}
\ELSE{
\STATE $\Delta \bar{t}_x \leftarrow |\Delta T_{x_-}|^{-1}\sum_{t_x \in \Delta T_{x_-}} t_x$
}
\ENDIF
\STATE Calculate $\Delta \bar{t}_y$ and $\Delta \bar{t}_z$ similarly.
\STATE $\Delta \bar{\mathbf{t}} \leftarrow (\Delta \bar{t}_x, \Delta \bar{t}_y, \Delta \bar{t}_z)$
\STATE Get the motion vector for the robot $\Delta \mathbf{t}_\text{robot}$.
\IF{$\cos\left<\Delta \bar{\mathbf{t}}, \Delta \mathbf{t}_\text{robot}\right> < 0$}{
\STATE $\lambda \leftarrow \lambda_o$
}
\ELSE{
\STATE $\lambda \leftarrow \lambda_p$
}
\ENDIF
\STATE $\bar{G}_j \leftarrow$ $\hat {G_j} + \lambda\Delta \bar{\mathbf{t}}$
\RETURN{$\bar{G}_j$}
\end{algorithmic}

\end{algorithm}

During the process of a dynamic handover, both the movement of the robot towards the grasp pose and the closing of the end-effector take an irreducible period of time. When the robot moves to the target pose and performs grasping, the object might change its position, leading to a handover failure. Therefore we add a simple but effective prediction module to predict the future grasp pose based on the existing dynamic grasp trajectory. In this module, we only consider the position prediction of the grasp pose since it is more important than rotation empirically.

The outline of the proposed future grasp prediction algorithm is depicted in Alg.~\ref{algo:alg1}. We utilize a grasp list $L$ to store the selected grasp pose in previous frames for the prediction of future grasp pose. Instead of simply calculating the ordinary moving momentum from the differential list, we adjust the moving momentum as follows for the robustness of the system against perturbations. First, we find out a relatively stable motion sequence and calculate the moving momentum during the selected motion (Line 3-5 in Alg.~\ref{algo:alg1}). Notice that here we allow small perturbations for longer sequences and better descriptions of motion patterns. Then, the motion direction along $x$-, $y$-, $z$-axis are determined by voting, \textit{i.e.}, following the majority of the motion direction along the axis in the stable motion sequence, and the momentum is adjusted using the determined direction (Line 6-12 in Alg.~\ref{algo:alg1}). Next, we adjust the coefficient of the momentum based on the idea that it takes more time for a robot to grasp an object that drives away from the robot than an object that moves towards the robot (Line 13-18 in Alg.~\ref{algo:alg1}). Finally, the adjusted moving momentum is added to the last grasp pose to predict the future grasp pose (Line 19 in Alg.~\ref{algo:alg1}).

In practice, we set the stability threshold $\delta_s$ for identifying stable sequence to $0.03\text{m}$ and the perturbation threshold $\delta_p$ to $0.005\text{m}$. The momentum coefficient for opposite motions $\lambda_o$ and parallel motions $\lambda_p$ are set to $1$ and $3$ respectively.

\section{System}

\begin{figure}[h]
    \centering
    
    \includegraphics[scale=0.7]{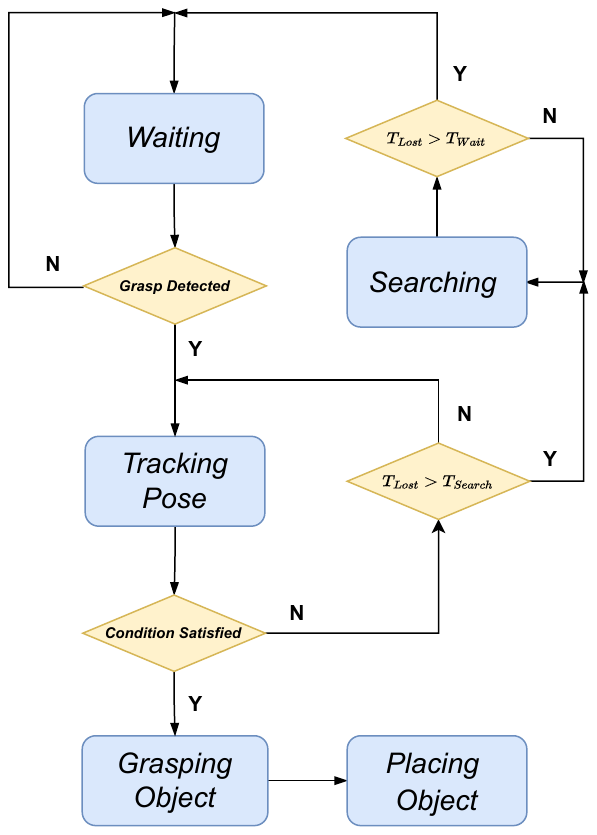}
    \caption{The pipeline of the closed-loop flexible handover system. There are five stages including waiting in the ready pose, tracking dynamic grasp, moving to the top pose to search, performing grasping and placing the object.}
    \label{fig:pipeline}
    
\end{figure}

The pipeline of our flexible handover system is shown in Fig.~\ref{fig:pipeline}. Given the first frame, we adopt the simplified GSNet~\cite{Wang_2021_ICCV} to detect the graspable candidates. After filtering out the unreasonable poses and passing the collision detection, the grasp pose with the highest score is selected as $\hat G_0$. We then utilize our tracking network to generate real-time dynamic grasp trajectory. Note that an extra hand segmentation model is not required during the tracking process, as our tracking network can track the grasp pose on an object. In the initial frame, a proper workspace is defined to discard the grasp poses detected on the hand. During the tracking stage, the robot moves to approach the predicted future pose $\bar G_j$ (Sec.~\ref{sec:prediction}). When the tracking pose in the current frame is lost, the tracking stage will be reinitialized, and the next frame will be considered as the first frame to detect $\hat G_0$. We denote the duration of frames with lost tracking as $T_{lost}$. If $T_{lost} > T_{search}$, which indicates that the object may have moved out of the current vision field, the robot will move to the top pose to obtain a wider field of view to detect new graspable poses. If $T_{lost} > T_{waiting}$, the handover process will be terminated, and the robot should return to the ready pose and wait. In practice, $T_{lost}$ and $T_{search}$ are set to 5 and 20 respectively.

We adopt online trajectory generator~\cite{kroger2009online} as motion planning. In the stage of tracking, we let the robot adapt the "top-down" pose to make the movement smoother and more flexible and move $\bar G_j$ along $z$-axis 3.5cm to avoid collision with the object during motion. When the predicted pose $\bar G_j$ is close enough to the robot's current pose, the robot will perform the grasping action.

\section{Experiments}

\subsection{Setup} 
\begin{figure}[htp]
    \centering

    \includegraphics[scale=0.32]{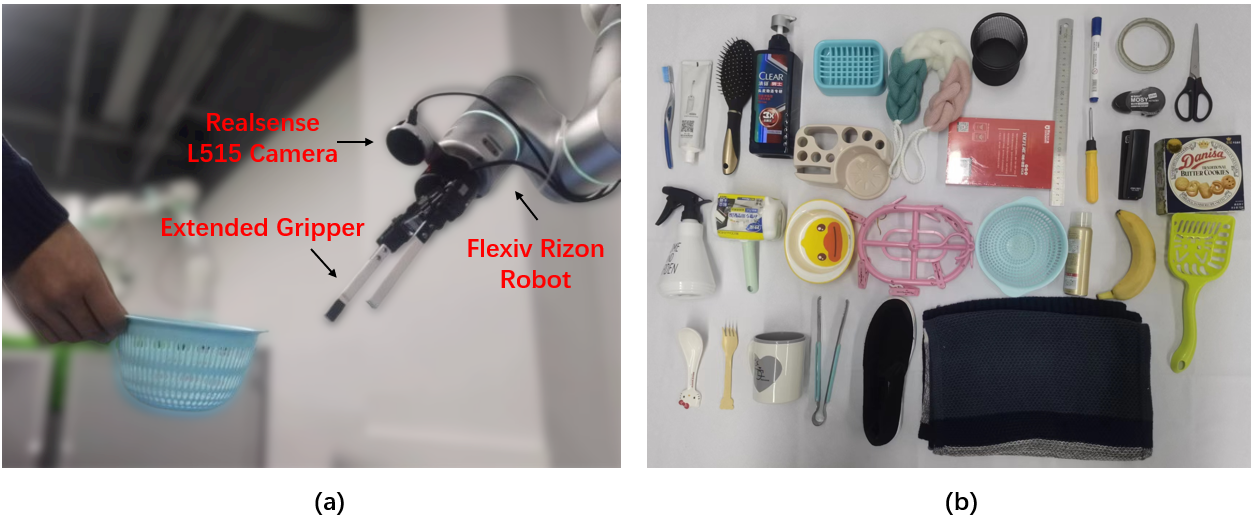}
    \caption{(a) The hardware setting or the handover system. An extended gripper is used to ensure that the depth camera can still produce accurate depth when moving close to the object. (b) The household object benchmark for dynamic handover experiments. The benchmark includes {toothbrush}, {toothpaste}, {comb}, {shampoo}, {soap box}, {cosmetic box}, {bath towel}, {pen holder}, {book}, {ruler}, {pen}, {screwdriver}, {tape}, {correction tape}, {stapler}, {scissors},  {paper box}, {spray bottle}, {lint roller}, {yellow basin}, {clothes horse}, {blue basket}, {bottle}, {banana}, {spatula}, {spoon}, {fork}, {mug}, {clip}, {shoe} and {scarf}.}
        \label{fig:settings}
\end{figure}

The hardware setting is shown in Fig.~\ref{fig:settings}(a). We use Flexiv Rizon as the robot. An Intel RealSense L515 camera is mounted on the robot. We use an extended Robotiq-85 gripper as the end-effector of the robot. The entire handover system runs on one NVIDIA RTX 3090 GPU with 6 FPS.

As illustrated in Fig.~\ref{fig:settings}(b), we have assembled a collection of 31 household objects for evaluation purposes. This diverse benchmark includes common household items, stationery objects, daily tools, food, and clothing, which are essential for manipulation tasks in a home robot scenario. The objects vary significantly in size, texture, color, material, and geometric structure. Our experiments are conducted using this benchmark.

\begin{table*}[bt]
    \setlength\tabcolsep{3pt}
    \centering
    
    \caption{Approaching time and success rate of handovers. We select six objects from the household object for our experiments. We rotate or translate the object after the robot begins to move. All the handovers are conducted until 3 successful trials.}
    \label{tab:one-motion}
    \begin{tabular}{l cc cc  cc}
    \toprule
         & \multicolumn{2}{c}{Banana} & \multicolumn{2}{c}{Bottle} & \multicolumn{2}{c}{Paper Box} \\
        Motion Pattern & Approaching Time (s) & Success Rate (\%) & Approaching Time (s) & Success Rate (\%) & Approaching Time (s) & Success Rate (\%)\\
    \midrule
         Rotation (0-$60^{\circ}$) &4.43 $\pm$ 1.86 &100\% & 4.44 $\pm$ 0.87 & 75\%&3.43 $\pm$ 0.35 & 100\%\\
         Translation (0-20cm) & 5.65 $\pm$ 0.18 &100\% & 5.42 $\pm$ 1.24 &100\% & 5.87 $\pm$ 0.49 & 100\% \\

    \toprule
     & \multicolumn{2}{c}{Mug} & \multicolumn{2}{c}{Pen} & \multicolumn{2}{c}{Spoon} \\
         Motion Pattern & Approaching Time (s) & Success Rate (\%) & Approaching Time (s) & Success Rate (\%) & Approaching Time (s) & Success Rate (\%)  \\
    \midrule
        Rotation (0-$60^{\circ}$) &5.29 $\pm$ 1.82 & 100\% &5.16 $\pm$ 1.36 &100\% &4.29 $\pm$ 1.38 &75\%\\
        Translation (0-20cm) & 5.77 $\pm$ 1.87 & 100\% & 5.18 $\pm$ 0.72 & 100\% & 5.86 $\pm$ 1.64 &75\%  \\
    \bottomrule
    \end{tabular}
\end{table*}

\begin{table*}[bt]
    \setlength\tabcolsep{3.2pt}
    \centering
    \caption{Success Rate(\%) of motion-continuous handover over the household object benchmark. Handovers for each object were performed until the target object was delivered successfully 3 times. }
    \label{tab:dynamic-handover}
    \begin{tabular}{cc cc cc cc cc cc cc cc cc}
    \toprule
     {  \textbf{Object}  } &{Toothbrush}&{Toothpaste}&{Comb} & {Shampoo} &{Soap Box} &{Cosmetic Box} &{Bath Towel}&{Pen Holder} \\
     \midrule
  {\textbf{Baseline / Ours}} & 100\% / 100\% & 100\% / 100\% & 50\% / 43\% & 43\% / 60\% &43\% / 100\%  & 100\% / 100\% & 60\% / 100\% & 75\% / 75\% \\
      \toprule
       {  \textbf{Object}  } &{ Book }&{Ruler} &{Pen} & {Tape} &{Screwdriver} & {Correction Tape} &{Stapler}&{Scissors}\\
          \midrule
        {\textbf{Baseline / Ours}} &50\% / 75\% & 43\% / 75\% & 50\% / 100\% & 50\% / 60 \% & 75\% / 100\% &75\% / 60\% &75\% / 100\% & 100\% / 75\% \\
      \toprule
           {  \textbf{Object}  } & {Paper Box} &{Spray bottle} & { Lint Roller} &{Yellow Basin} & {Clothes Horse}& {Blue Basket}& {Bottle}&{Banana}\\
          \midrule
    {\textbf{Baseline / Ours}} & 50\% / 60\% & 43\% / 100\% & 50\% / 75\% & 75\% / 100\% & 75\% / 75\% & 50\% / 75\%& 60\% / 100\% & 50\% / 60\%  \\
      \toprule
           {  \textbf{Object}  } & {Spatula} &{Spoon}& {Mug}& {Fork}&{Clip}&{Shoe}&{Scarf }&{\textbf{Overall}}\\
          \midrule
       {\textbf{Baseline / Ours}} &100\% / 75\% &43\% / 75\% & 75\% / 75\% &75\% / 100\% &75\% / 75\%& 50\% / 75\% &100\% / 100\% & \textbf{60.78\%} / \textbf{78.15\%}  \\
      \bottomrule

    \end{tabular}
\end{table*}

\subsection{One-Motion Handover Experiments}

\begin{figure*}[t!]
    \centering
    \includegraphics[scale=0.6]{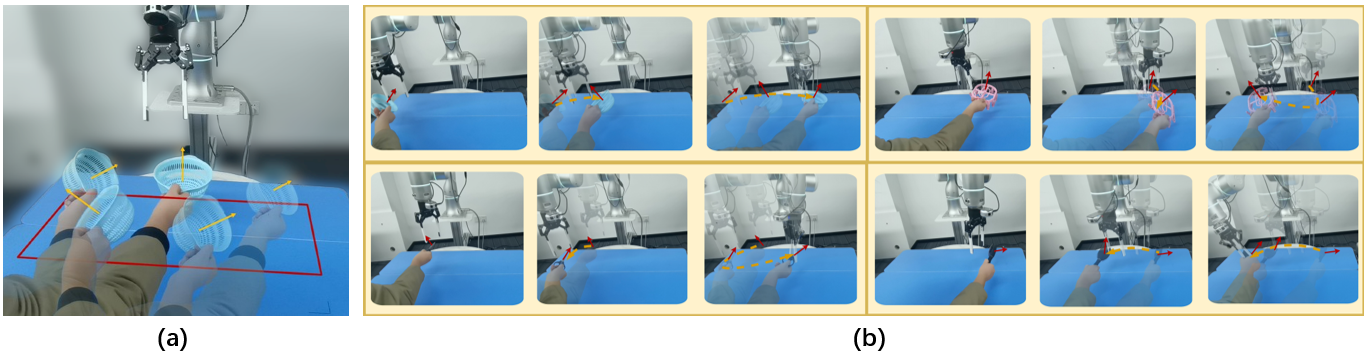}
    \caption{(a) The setup of motion-continuous handover experiments. The red line denotes the range of translation. The yellow arrow denotes the rotation in motion. (b) Some qualitative results of our flexible handover experiments. The trajectory of moving objects and the change of rotation are are shown in the pictures. The objects in the picture are a blue basket, a clothes horse, scissors, and a comb respectively.}
    \label{fig:dynamic-experiment}
\end{figure*}

The one-motion handover experiment mainly follows~\cite{yang2021reactive}. We select 6 objects from the household object benchmark. First, we select banana, bottle, and paper box that are similar to the objects in~\cite{yang2021reactive}'s systematic evaluation. Then, we select the most difficult three objects in~\cite{yang2021reactive}'s user study: mug, pen, and spoon, whose success rates are all below 70\% in~\cite{yang2021reactive}. We move the object once following different motion patterns after the robot begins to move. Table~\ref{tab:one-motion} shows the approach time and success rate over the six objects under different motion patterns (rotation and translation). The one-motion handover is conducted until the robot receives the object in hand successfully 3 times. We can conclude that the average time for handover is 5.07s and the overall success rate is 92.3\%. Our system performs well in terms of both success rate and speed.

\subsection{Motion-Continuous Handover Experiments}
We conduct motion-continuous handover experiments to verify the effectiveness and robustness of our flexible handover system under situations where the object in hand has a larger range of movement and more complex motion trajectories (including both transition and rotation). The experiment setup is shown in Fig.~\ref{fig:dynamic-experiment}(a). Taking the center of the workspace as the origin, the translation range on the xOy plane is $[-0.3\text{m},0.3\text{m}]\times [-0.15\text{m},0.15\text{m}]$; and the orientation range is $\left[-60^{\circ}, 60^{\circ}\right]$. The experiment participant can move the object in hand inside the range at a reasonable speed, and then the robot will try to approach and grasp the moving object. Some qualitative results of the experiments are visualized in Fig.~\ref{fig:dynamic-experiment}(b).

We employ a baseline that chooses the nearest pose with the tracking pose in the last frame~\cite{morrison2018closing} in the experiments. For comparison, the experimenter moves the objects to be handed over following a similar pattern and speed in the experiment utilizing the two different methods. Table \ref{tab:dynamic-handover} shows the success rate of different objects with the baseline or our method over the household benchmark. The overall success rate of our proposed flexible handover system is $78.15\%$, outperforming the success rate of $60.78\%$ achieved by the baseline, which shows the superiority of our proposed method. For the baseline method, we find that more failure cases are caused by grasping positions with bad quality because the consistency of grasp quality is not guaranteed under complex motion. We also observe that the baseline method exhibits a slower reaction speed to approach moving objects, in contrast to our proposed method.

As to the comparison with previous work, Wang \textit{et al.}~\cite{wang2022goal} improved \cite{yang2021reactive}'s handover system by employing their closed-loop grasping policy. They performed the dynamic handover experiment where the object is moved from left to right (L2R) and from right to left (R2L) during the entire handover on a smaller benchmark including 10 household objects based on \cite{yang2021reactive}'s perception system. 6 out of 10 objects in their benchmark are similar or identical to those in our benchmark. Since their handover system is not fully publicly available, we cannot reproduce and compare to their results directly. Over the similar object list, they reported success rates of $83.33\%$ and $50.00\%$ in L2R and R2L experiments respectively in their original paper. In contrast, we achieved a success rate of $78.26\%$ in our proposed motion-continuous handover experiment, which involves more complex motion patterns of the objects.

\section{Conclusions}

We develop a general, robust, and efficient flexible handover system that incorporates general grasp pose detection, real-time robust grasp trajectory generation, and future grasp prediction. Our system enables users to move and rotate objects within a large range for the robot to approach, track, and grasp. We have demonstrated that our system exhibits high reactivity and success rates. We believe that this flexible handover system can serve as a foundation for more complex collaborative tasks in the future.

\section*{Acknowledgments}
This work is supported in part by the National Key R\&D Program of China, No. 2017YFA0700800, Shanghai Municipal Science
and Technology Major Project (2021SHZDZX0102), Shanghai Qi Zhi Institute, SHEITC (2018-RGZN-02046). 

\small {\textit{Authors' contributions}: H.-S. Fang initiated the project with G.~Zhang. G. Zhang implemented the grasp tracking algorithm, ran the experiments, and wrote the paper. H.-S. Fang mentored this project and edited the paper. H. Fang implemented the robot closed-loop control. C. Lu supervised the project and provided the environment and hardware support for this research. }

\bibliographystyle{IEEEtranS}
\bibliography{reference}

\begin{thebibliography}{10}
\providecommand{\url}[1]{#1}
\csname url@rmstyle\endcsname
\providecommand{\newblock}{\relax}
\providecommand{\bibinfo}[2]{#2}
\providecommand\BIBentrySTDinterwordspacing{\spaceskip=0pt\relax}
\providecommand\BIBentryALTinterwordstretchfactor{4}
\providecommand\BIBentryALTinterwordspacing{\spaceskip=\fontdimen2\font plus
\BIBentryALTinterwordstretchfactor\fontdimen3\font minus
  \fontdimen4\font\relax}
\providecommand\BIBforeignlanguage[2]{{%
\expandafter\ifx\csname l@#1\endcsname\relax
\typeout{** WARNING: IEEEtran.bst: No hyphenation pattern has been}%
\typeout{** loaded for the language `#1'. Using the pattern for}%
\typeout{** the default language instead.}%
\else
\language=\csname l@#1\endcsname
\fi
#2}}

\bibitem{ortenzi2021object}
V.~Ortenzi, A.~Cosgun, T.~Pardi, W.~P. Chan, E.~Croft, and D.~Kuli{'c},
  ``Object handovers: a review for robotics,'' \emph{IEEE Transactions on
  Robotics}, vol.~37, no.~6, pp. 1855--1873, 2021.

\bibitem{rosenberger2020object}
P.~Rosenberger, A.~Cosgun, R.~Newbury, J.~Kwan, V.~Ortenzi, P.~Corke, and
  M.~Grafinger, ``Object-independent human-to-robot handovers using real time
  robotic vision,'' \emph{IEEE Robotics and Automation Letters}, vol.~6, no.~1,
  pp. 17--23, 2020.

\bibitem{yang2021reactive}
W.~Yang, C.~Paxton, A.~Mousavian, Y.-W. Chao, M.~Cakmak, and D.~Fox, ``Reactive
  human-to-robot handovers of arbitrary objects,'' in \emph{2021 IEEE
  International Conference on Robotics and Automation (ICRA)}.\hskip 1em plus
  0.5em minus 0.4em\relax IEEE, 2021, pp. 3118--3124.

\bibitem{marturi2019dynamic}
N.~Marturi, M.~Kopicki, A.~Rastegarpanah, V.~Rajasekaran, M.~Adjigble,
  R.~Stolkin, A.~Leonardis, and Y.~Bekiroglu, ``Dynamic grasp and trajectory
  planning for moving objects,'' \emph{Autonomous Robots}, vol.~43, pp.
  1241--1256, 2019.

\bibitem{mavsar2022rovernet}
M.~Mavsar and A.~Ude, ``Rovernet: Vision-based adaptive human-to-robot object
  handovers,'' in \emph{2022 IEEE-RAS 21st International Conference on Humanoid
  Robots (Humanoids)}.\hskip 1em plus 0.5em minus 0.4em\relax IEEE, 2022, pp.
  858--864.

\bibitem{Wang_2021_ICCV}
C.~Wang, H.-S. Fang, M.~Gou, H.~Fang, J.~Gao, and C.~Lu, ``Graspness discovery
  in clutters for fast and accurate grasp detection,'' in \emph{Proceedings of
  the IEEE/CVF International Conference on Computer Vision (ICCV)}, October
  2021, pp. 15\,964--15\,973.

\bibitem{strabala2013toward}
K.~Strabala, M.~K. Lee, A.~Dragan, J.~Forlizzi, S.~S. Srinivasa, M.~Cakmak, and
  V.~Micelli, ``Toward seamless human-robot handovers,'' \emph{Journal of
  Human-Robot Interaction}, vol.~2, no.~1, pp. 112--132, 2013.

\bibitem{aleotti2012comfortable}
J.~Aleotti, V.~Micelli, and S.~Caselli, ``Comfortable robot to human object
  hand-over,'' in \emph{2012 IEEE RO-MAN: The 21st IEEE International Symposium
  on Robot and Human Interactive Communication}.\hskip 1em plus 0.5em minus
  0.4em\relax IEEE, 2012, pp. 771--776.

\bibitem{grigore2013joint}
E.~C. Grigore, K.~Eder, A.~G. Pipe, C.~Melhuish, and U.~Leonards, ``Joint
  action understanding improves robot-to-human object handover,'' in \emph{2013
  IEEE/RSJ international conference on intelligent robots and systems}.\hskip
  1em plus 0.5em minus 0.4em\relax IEEE, 2013, pp. 4622--4629.

\bibitem{nagata1998delivery}
K.~Nagata, Y.~Oosaki, M.~Kakikura, and H.~Tsukune, ``Delivery by hand between
  human and robot based on fingertip force-torque information,'' in
  \emph{Proceedings. 1998 IEEE/RSJ International Conference on Intelligent
  Robots and Systems. Innovations in Theory, Practice and Applications (Cat.
  No. 98CH36190)}, vol.~2.\hskip 1em plus 0.5em minus 0.4em\relax IEEE, 1998,
  pp. 750--757.

\bibitem{edsinger2007human}
A.~Edsinger and C.~C. Kemp, ``Human-robot interaction for cooperative
  manipulation: Handing objects to one another,'' in \emph{RO-MAN 2007-The 16th
  IEEE International Symposium on Robot and Human Interactive
  Communication}.\hskip 1em plus 0.5em minus 0.4em\relax IEEE, 2007, pp.
  1167--1172.

\bibitem{wang2018controlling}
W.~Wang, R.~Li, Z.~M. Diekel, Y.~Chen, Z.~Zhang, and Y.~Jia, ``Controlling
  object hand-over in human--robot collaboration via natural wearable
  sensing,'' \emph{IEEE Transactions on Human-Machine Systems}, vol.~49, no.~1,
  pp. 59--71, 2018.

\bibitem{kajikawa2000trajectory}
S.~Kajikawa and E.~Ishikawa, ``Trajectory planning for hand-over between human
  and robot,'' in \emph{Proceedings 9th IEEE International Workshop on Robot
  and Human Interactive Communication. IEEE RO-MAN 2000 (Cat. No.
  00TH8499)}.\hskip 1em plus 0.5em minus 0.4em\relax IEEE, 2000, pp. 281--287.

\bibitem{vogt2018one}
D.~Vogt, S.~Stepputtis, B.~Jung, and H.~B. Amor, ``One-shot learning of
  human--robot handovers with triadic interaction meshes,'' \emph{Autonomous
  Robots}, vol.~42, pp. 1053--1065, 2018.

\bibitem{yang2022model}
W.~Yang, B.~Sundaralingam, C.~Paxton, I.~Akinola, Y.-W. Chao, M.~Cakmak, and
  D.~Fox, ``Model predictive control for fluid human-to-robot handovers,'' in
  \emph{2022 International Conference on Robotics and Automation (ICRA)}.\hskip
  1em plus 0.5em minus 0.4em\relax IEEE, 2022, pp. 6956--6962.

\bibitem{prada2014implementation}
M.~Prada, A.~Remazeilles, A.~Koene, and S.~Endo, ``Implementation and
  experimental validation of dynamic movement primitives for object handover,''
  in \emph{2014 IEEE/RSJ International Conference on Intelligent Robots and
  Systems}.\hskip 1em plus 0.5em minus 0.4em\relax IEEE, 2014, pp. 2146--2153.

\bibitem{bianchi2018touch}
M.~Bianchi, G.~Averta, E.~Battaglia, C.~Rosales, M.~Bonilla, A.~Tondo,
  M.~Poggiani, G.~Santaera, S.~Ciotti, M.~G. Catalano, \emph{et~al.},
  ``Touch-based grasp primitives for soft hands: Applications to human-to-robot
  handover tasks and beyond,'' in \emph{2018 IEEE International Conference on
  Robotics and Automation (ICRA)}.\hskip 1em plus 0.5em minus 0.4em\relax IEEE,
  2018, pp. 7794--7801.

\bibitem{konstantinova2017autonomous}
J.~Konstantinova, S.~Krivic, A.~Stilli, J.~Piater, and K.~Althoefer,
  ``Autonomous object handover using wrist tactile information,'' in
  \emph{Towards Autonomous Robotic Systems: 18th Annual Conference, TAROS 2017,
  Guildford, UK, July 19--21, 2017, Proceedings 18}.\hskip 1em plus 0.5em minus
  0.4em\relax Springer, 2017, pp. 450--463.

\bibitem{mazhitov2023human}
A.~Mazhitov, T.~Syrymova, Z.~Kappassov, and M.~Rubagotti, ``Human--robot
  handover with prior-to-pass soft/rigid object classification via tactile
  glove,'' \emph{Robotics and Autonomous Systems}, vol. 159, p. 104311, 2023.

\bibitem{redmon2018yolov3}
J.~Redmon and A.~Farhadi, ``Yolov3: An incremental improvement,'' \emph{arXiv
  preprint arXiv:1804.02767}, 2018.

\bibitem{morrison2018closing}
D.~Morrison, J.~Leitner, and P.~Corke, ``Closing the loop for robotic grasping:
  {A} real-time, generative grasp synthesis approach,'' in \emph{Robotics:
  Science and Systems}, 2018.

\bibitem{mousavian20196}
A.~Mousavian, C.~Eppner, and D.~Fox, ``6-dof graspnet: Variational grasp
  generation for object manipulation,'' in \emph{Proceedings of the IEEE/CVF
  International Conference on Computer Vision}, 2019, pp. 2901--2910.

\bibitem{aitor2012point}
A.~Aitor, Z.~Marton, F.~Tombari, W.~Wohlkinger, C.~Potthast, B.~Zeisl, R.~Rusu,
  S.~Gedikli, and M.~Vincze, ``Point cloud library,'' \emph{IEEE Robotics \&
  Automation Magazine 1070.9932/12}, 2012.

\bibitem{fang2023anygrasp}
H.-S. Fang, C.~Wang, H.~Fang, M.~Gou, J.~Liu, H.~Yan, W.~Liu, Y.~Xie, and
  C.~Lu, ``Anygrasp: Robust and efficient grasp perception in spatial and
  temporal domains,'' \emph{IEEE Transactions on Robotics (T-RO)}, 2023.

\bibitem{Liu_2023_CVPR}
J.~Liu, R.~Zhang, H.-S. Fang, M.~Gou, H.~Fang, C.~Wang, S.~Xu, H.~Yan, and
  C.~Lu, ``Target-referenced reactive grasping for dynamic objects,'' in
  \emph{Proceedings of the IEEE/CVF Conference on Computer Vision and Pattern
  Recognition (CVPR)}, June 2023, pp. 8824--8833.

\bibitem{wang2022goal}
L.~Wang, Y.~Xiang, W.~Yang, A.~Mousavian, and D.~Fox, ``Goal-auxiliary
  actor-critic for 6d robotic grasping with point clouds,'' in \emph{Conference
  on Robot Learning}.\hskip 1em plus 0.5em minus 0.4em\relax PMLR, 2022, pp.
  70--80.

\bibitem{diakogiannis2020resunet}
F.~I. Diakogiannis, F.~Waldner, P.~Caccetta, and C.~Wu, ``Resunet-a: A deep
  learning framework for semantic segmentation of remotely sensed data,''
  \emph{ISPRS Journal of Photogrammetry and Remote Sensing}, vol. 162, pp.
  94--114, 2020.

\bibitem{choy20194d}
C.~Choy, J.~Gwak, and S.~Savarese, ``4d spatio-temporal convnets: Minkowski
  convolutional neural networks,'' in \emph{Proceedings of the IEEE/CVF
  Conference on Computer Vision and Pattern Recognition}, 2019, pp. 3075--3084.

\bibitem{vaswani2017attention}
A.~Vaswani, N.~Shazeer, N.~Parmar, J.~Uszkoreit, L.~Jones, A.~N. Gomez,
  {\L}.~Kaiser, and I.~Polosukhin, ``Attention is all you need,''
  \emph{Advances in neural information processing systems}, vol.~30, 2017.

\bibitem{fang2020graspnet}
H.-S. Fang, C.~Wang, M.~Gou, and C.~Lu, ``Graspnet-1billion: A large-scale
  benchmark for general object grasping,'' in \emph{Proceedings of the IEEE/CVF
  Conference on Computer Vision and Pattern Recognition}, 2020, pp.
  11\,444--11\,453.

\bibitem{gou2021RGB}
M.~Gou, H.-S. Fang, Z.~Zhu, S.~Xu, C.~Wang, and C.~Lu, ``Rgb matters: Learning
  7-dof grasp poses on monocular rgbd images,'' in \emph{Proceedings of the
  International Conference on Robotics and Automation (ICRA)}, 2021.

\bibitem{zhou2022global}
X.~Zhou, T.~Yin, V.~Koltun, and P.~Kr{\"a}henb{\"u}hl, ``Global tracking
  transformers,'' in \emph{Proceedings of the IEEE/CVF Conference on Computer
  Vision and Pattern Recognition}, 2022, pp. 8771--8780.

\bibitem{glorot2011deep}
X.~Glorot, A.~Bordes, and Y.~Bengio, ``Deep sparse rectifier neural networks,''
  in \emph{Proceedings of the fourteenth international conference on artificial
  intelligence and statistics}.\hskip 1em plus 0.5em minus 0.4em\relax JMLR
  Workshop and Conference Proceedings, 2011, pp. 315--323.

\bibitem{hendrycks2016bridging}
D.~Hendrycks and K.~Gimpel, ``Bridging nonlinearities and stochastic
  regularizers with gaussian error linear units,'' \emph{CoRR, abs/1606.08415},
  vol.~3, 2016.

\bibitem{srivastava2014dropout}
N.~Srivastava, G.~Hinton, A.~Krizhevsky, I.~Sutskever, and R.~Salakhutdinov,
  ``Dropout: a simple way to prevent neural networks from overfitting,''
  \emph{The journal of machine learning research}, vol.~15, no.~1, pp.
  1929--1958, 2014.

\bibitem{kingma2014adam}
D.~P. Kingma and J.~Ba, ``Adam: A method for stochastic optimization,''
  \emph{arXiv preprint arXiv:1412.6980}, 2014.

\bibitem{chen2017deeplab}
L.-C. Chen, G.~Papandreou, I.~Kokkinos, K.~Murphy, and A.~L. Yuille, ``Deeplab:
  Semantic image segmentation with deep convolutional nets, atrous convolution,
  and fully connected crfs,'' \emph{IEEE transactions on pattern analysis and
  machine intelligence}, vol.~40, no.~4, pp. 834--848, 2017.

\bibitem{kroger2009online}
T.~Kr{\"o}ger and F.~M. Wahl, ``Online trajectory generation: Basic concepts
  for instantaneous reactions to unforeseen events,'' \emph{IEEE Transactions
  on Robotics}, vol.~26, no.~1, pp. 94--111, 2009.

\end{thebibliography}

\end{document}